\documentclass[journal]{IEEEtran}
\usepackage{graphicx}
\usepackage{subfigure}  
\usepackage{caption}
\usepackage{float}
\usepackage{multirow} 
\usepackage{color}
\usepackage{longtable} 
\usepackage{array} 
\usepackage{makecell}
\usepackage{threeparttable} 
\usepackage{booktabs}
\usepackage{lineno,hyperref}
\modulolinenumbers[5]
\usepackage{multirow}
\usepackage{booktabs}
\usepackage{amsmath,amssymb}
\usepackage{fancyhdr}
\ifCLASSINFOpdf
\else
\fi
\begin{document}
\captionsetup[figure]{labelformat={default},labelsep=period,name={Fig.}}
%
\title{Visualizing Deep Learning-based Radio Modulation Classifier}
%
%
%

\author{Liang~Huang,~\IEEEmembership{Member,~IEEE,}
        You~Zhang,
        Weijian~Pan,
        Jinyin~Chen,
        Li~Ping~Qian,~\IEEEmembership{Senior~Member,~IEEE}
        and Yuan~Wu,~\IEEEmembership{Senior~Member,~IEEE}
\thanks{L. Huang is with the College of Computer Science and Technology, Y. Zhang, W. Pan, and L. P. Qian are with the College of Information Engineering, J. Chen is with the Institute of Cyberspace Security, Zhejiang University of Technology, Hangzhou, China 310023. L. P. Qian is also with the National Mobile Communications Research Laboratory, Southeast University, Nanjing 210096, China (e-mail: \{lianghuang, 2111803267, 2111803213, lpqian, chenjinyin\}@zjut.edu.cn). (Corresponding author: Jinyin Chen.)}

\thanks{Y. Wu is with State Key Laboratory of Internet of Things for Smart City, University of Macau, Macao, China. (e-mail: yuanwu@um.edu.mo).}
}

\maketitle
\thispagestyle{fancy}
\begin{abstract}\label{abstract}
Deep learning has recently been successfully applied in automatic modulation classification by extracting and classifying radio features in an end-to-end way. However, deep learning-based radio modulation classifiers are lacking interpretability, and there is little explanation or visibility into what kinds of radio features are extracted and chosen for classification. In this paper, we visualize different deep learning-based radio modulation classifiers by introducing a class activation vector. Specifically, both convolutional neural networks (CNN) based classifier and long short-term memory (LSTM) based classifier are separately studied, and their extracted radio features are visualized. We explore different hyperparameter settings via extensive numerical evaluations and show both the CNN-based classifier and LSTM-based classifiers extract similar radio features relating to modulation reference points. In particular, for the LSTM-based classifier, its obtained radio features are similar to the knowledge  of human experts. Our numerical results indicate the radio features extracted by deep learning-based classifiers greatly depend on the contents carried by radio signals, and a short radio sample may lead to misclassification.
\end{abstract}

%
\IEEEpeerreviewmaketitle

\section{Introduction}\label{section:introduction}
%
%
%
%
\IEEEPARstart{A}{utomatic} modulation classification (AMC) detects the modulation categories of received signals for further demodulation, which plays an important role in civilian and military applications \cite{zhu2015automatic}, i.e., cognitive radio, software-defined radio, and electronic warfare. Traditional AMC methods first extract radio features based on expert knowledge i.e., spectrum \cite{nandi1995automatic}, moments \cite{soliman1992signal}, cumulants \cite{swami2000hierarchical}\cite{majhi2017hierarchical}, and then classify them via statistical or machine learning algorithms. However, the classification accuracy greatly depends on the extracted radio-specific features whose performance cannot be guaranteed. It is challenging and compute-intensive to extract the right radio features. Recently, deep learning-based radio modulation classifiers \cite{10.1007/978-3-319-44188-7_16}\cite{rajendran2018deep} are proposed to conduct feature extraction and classification at the same time via deep neural network (DNN). They use raw radio signals as DNN input, automatically learn radio features with multiple levels of abstraction \cite{lecun2015deep} hidden in the data, and achieve significant improvements in classification accuracy.

Although deep learning has been successfully applied to AMC, it is still an open question regarding the understanding and the key reason about the modulation classification  mechanism, namely the interpretability \cite{8397411}. Different from traditional highly interpretable feature-based AMC algorithms, deep learning-based classifiers infer the modulation categories in an end-to-end way, operating as ``black boxes''. In recent years, the interpretability of deep learning-based classifiers has been gradually studied in the fields of image classification \cite{7234886}, natural language processing \cite{10.1145/1390156.1390177}, speech recognition \cite{hinton2012deep}, and text classification \cite{10.1145/3077136.3080834}. However, in the field of radio signals, there is little explanation or visibility into what kinds of radio features are extracted by different deep learning-based radio modulation classifiers. 

In this paper, we study visualization methods for deep learning-based radio modulation classifiers. Specifically, two state-of-the-art modulation classifiers based on convolutional neural networks (CNN) and long short-term memory (LSTM) are studied and their extracted radio features are visualized. After extensive evaluations on an open radio signal dataset, we obtain the following contributions:
\begin{itemize}
	\item We present a visualization structure based on a class activation vector for different deep learning-based radio modulation classifiers. Each element value of the class activation vector represents the significance of the corresponding radio signal sample point in modulation classification. By introducing an activation threshold, we further visualize the time-domain radio features by connecting those consecutive sample points whose corresponding element values are greater than the threshold.
	\item We visualize that both CNN-based and LSTM-based classifiers extract similar radio features for the same modulation category. However, the CNN-based classifier captures the radio signal transitions from one modulation reference point to another. On the other hand, the LSTM-based classifier only works with radio signals in the amplitude/phase format and focuses on those sample points close to the modulation reference points, which is similar to the knowledge of human experts.
	\item We further evaluate radio signals with fewer sample points via the ResNet-based classifier. We visually illustrate that the radio features extracted by the deep learning-based classifier greatly depend on the contents carried by radio signals and a short radio sample may lead to misclassification.
\end{itemize}

The remainder of this paper is organized as follows. We describe the related works on deep learning-based radio modulation classification and visualization in Section II. In Section III, we provide an overview of deep learning-based modulation classifiers. In Section IV, we propose a visualization scheme based on a class activation vector and visualize both CNN-based and LSTM-based radio modulation classifiers. In Section V, we present numerical results. This paper is concluded in Section VI.

\section{Related works}\label{2}
\subsection{Deep Learning-based Radio Modulation Classification}
 Deep learning-based classifiers have been successfully applied to automatically classify radio modulation categories in recent literature. For example, \cite{10.1007/978-3-319-44188-7_16} proposed LeNet-based modulation classifier which uses modulated in-phase and quadrature-phase signals as neural networks' input. By considering wireless channels with impacts of multipath fading, sample rate offset, and center frequency offset, they show that the LeNet-based classifier outperforms expert features-based algorithms, especially for radio signals with low SNR and short sample. In \cite{o2018over} the authors further improved the performance by using a ResNet-based classifier. By using radio signals' amplitude and phase information as the network input, \cite{rajendran2018deep} proved an LSTM-based modulation classifier which outperforms the existing CNN-based algorithm. \cite{app10031166} proposed a modulation classification algorithm combining the InceptionResNetV2 network with transfer adaptation to further improve the classification accuracy. \cite{chen2020novel} proposed an attention cooperative framework to improve the classification accuracy and \cite{8949478} exploited the graph convolutional network. Moreover, other works transformed 
the radio signals into images, i.e., constellation diagram \cite{peng2018modulation}, spectrogram \cite{8587447}, and 
classified the modulation categories using existing image classifiers. Furthermore, different data augmentation methods are studied \cite{8936957}\cite{antoniou2017data} to better train deep learning-based classifiers.

\subsection{Visualization of Deep Learning-based Classifier}
With the continuous development of deep learning technology, several visualization technologies have been proposed. Researchers are interested in exploring the decision mechanism inside the black box. Extensive visualization technologies for image classifiers have been successfully proposed, i.e., activation maximization \cite{simonyan2014a,mahendran2016visualizing}, variants of deconvolution and back propagation \cite{DB15a,10.1007/978-3-319-10590-1_53}, network inversion \cite{Mahendran_2015_CVPR}, and feature area \cite{Selvaraju_2017_ICCV,Fong_2017_ICCV}. Besides image processing, \cite{liu2017visualization} studied the visualization of driving behavior feature extraction based on deep learning. In the medical field, visualization techniques are used to explain how decisions are made for various deep models of electroencephalographic data \cite{schirrmeister2017deep}. In the field of biology, \cite{lanchantin2017deep} visualized DNNs and studied how the neural network makes decisions in predicting the transcription factor binding site tasks. In the field of speech recognition, \cite{7952654,7472084} studied the mechanism behind the outstanding performance of recurrent neural networks in processing speech through visualization techniques. In Natural language processing, \cite{NIPS2013_5166,karpathy2015visualizing} proposed a variety of different visualization methods to help people understand how recurrent neural networks make decisions, such as building sentences based on the meaning of words and phrases. Although deep learning models are widely used in the field of radio modulation classification, their classification mechanism is still unclear. To the best of our knowledge, visualization techniques are not used in this field. In this paper, we study the visualization methods for deep learning-based radio modulation classifiers. 
%
%
\section{Deep Learning-based Radio Modulation Classifier}\label{3}
Consider a segment of sampled radio signal with length $N_x$, $\boldsymbol{x}=\{x_i| i\in\mathbb{N}_x\}$ where $\mathbb{N}_x=\{0,1,2,\dots,N_x-1\}$, which belongs to one of $N_y$ different modulation categories, denoted by a labeled set $\mathbb{N}_y=\{0,1,2,\dots,N_y-1\}$. A radio modulation classifier maps $\boldsymbol{x}$ to a vector $\boldsymbol{y}=\{y_j\in(0,1)|j\in\mathbb{N}_y\}$, where $y_j$ denotes the probability that the segmented signal $\boldsymbol{x}$ belongs to the j-th modulation category. The mapping function can be denoted as
\begin{center}
	$\pi$: $\boldsymbol{x}\mapsto\boldsymbol{y}$.
\end{center}
Then, the predicted modulation category is the one with largest probability $y_j$, denoted as $j^{*}=\arg \max _{j \in \mathbb{N}_{y}} y_{j}$.

Different deep neural network models based on CNN \cite{10.1007/978-3-319-44188-7_16,o2018over} or LSTM \cite{rajendran2018deep} have been developed to successfully classify radio modulation categories. In general, they use raw sampled data as the input of the classifiers, i.e., the modulated in-phase (I) and quadrature-phase (Q) samples $x_{i}=\left(I_{i}, Q_{i}\right)$  or the transformed amplitude ($\mathrm{A}$) and phase ($\phi$) samples $x_{i}=\left(\mathrm{A}_{i}, \phi_{i}\right)$ via 
\begin{eqnarray}
\left\{\begin{array}{l}
A_{i}=\sqrt{I_{i}^{2}+Q_{i}^{2}} \\
\phi_{i}=\arctan \left(Q_{i} / I_{i}\right)
\end{array}\right..
\end{eqnarray}
Without extracting expert features, deep learning-based modulation classifiers directly output the predicted probabilities $\boldsymbol{y}$ and achieve significant classification accuracy. However, the raw data provides little insight on how these deep learning-based models classify radio modulation categories.

\section{Visualization of Deep Learning-based Radio Modulation Classifier}\label{4}
\begin{figure}
	\centering
	\includegraphics[scale=0.9]{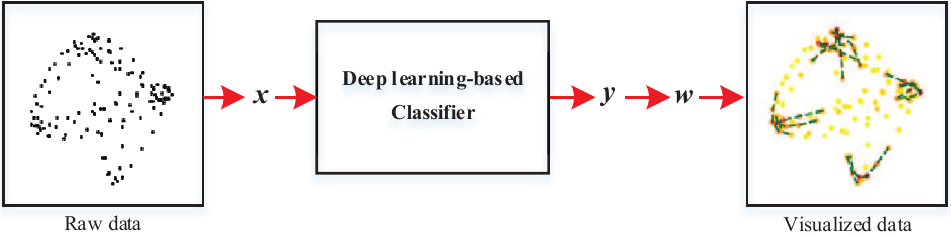}
	\caption{The schematics of visualizing deep learning-based radio modulation classifier.}
	\label{fig:1}
\end{figure}

We aim to visualize the classifier by introducing a class activation vector $\boldsymbol{w}=\left\{w_{i} \in[0,1] | i \in \mathbb{N}_{x}\right\}$, where each $w_{i}$  represents the significance of input $x_i$ on classifying the modulation category. The visualization function is defined as
\begin{center}
	$g_{\pi}:(\boldsymbol{x}, \boldsymbol{y}) \mapsto \boldsymbol{w}$.
\end{center}
\begin{figure*}
	\centering
	\includegraphics[scale=1]{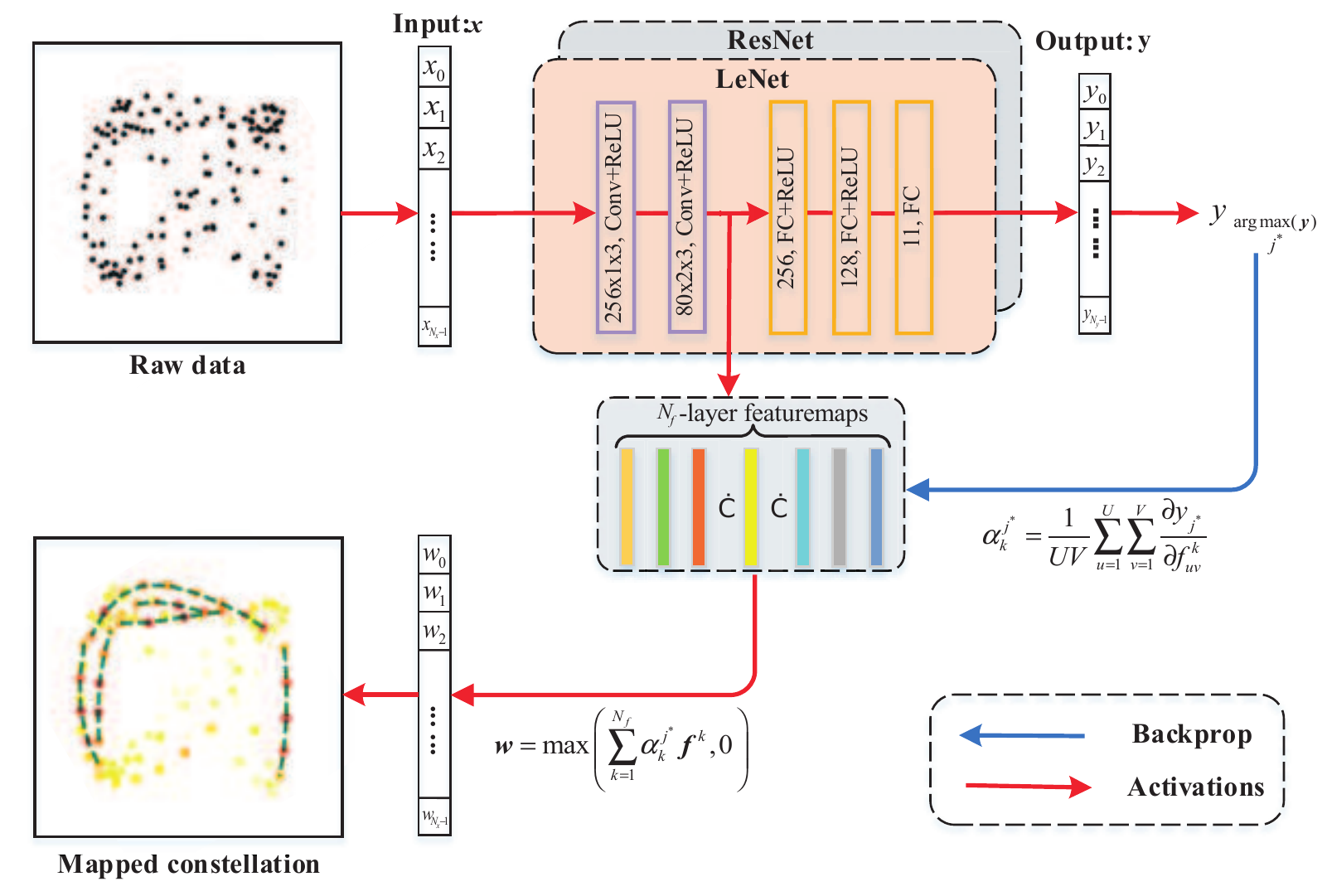}
	\caption{The schematics of visualizing CNN-based radio modulation classifier.}
	\label{fig:2}
\end{figure*}

\subsection{Visualization Overview}\label{4.1}
The structure of the visualization method is illustrated in Fig.~\ref{fig:1}. For better illustration, we plot the constellation diagram of the consecutive radio signal samples $\boldsymbol{x}$. Each constellation point corresponds to a radio signal sample $x_i$ plotted in the complex plane in the Cartesian coordinate system $x_{i}=\left(I_{i}, Q_{i}\right)$ or the polar coordinate system $x_{i}=\left(\mathrm{A}_{i}, \phi_{i}\right)$. Then, we color each constellation point ${x}_{i}$ with a unique color ranging from yellow to red depending on the obtain weight ${w}_{i}$. A sample point with large weight ${w}_{i}$ (close to 1) is colored in red, meaning that it is important for classifying the modulation category. 

To visualize the time-domain feature, we further introduce an activation threshold $\eta_{w}$ and connect each pair of two consecutive sample points ${x}_{i}$ and ${x}_{i+1}$ via a green line when both their weights are greater than the threshold $\eta_{w}$, as $w_{i}$, $w_{i+1}>\eta_{w}$.  In the following subsections, we evaluate the class activation vectors for both CNN-based and LSTM-based classifiers by adopting the Grad-CAM \cite{Selvaraju_2017_ICCV} algorithm and the mask-based algorithm \cite{Fong_2017_ICCV}, respectively.
\begin{figure}
	\centering
	\includegraphics[scale=0.9]{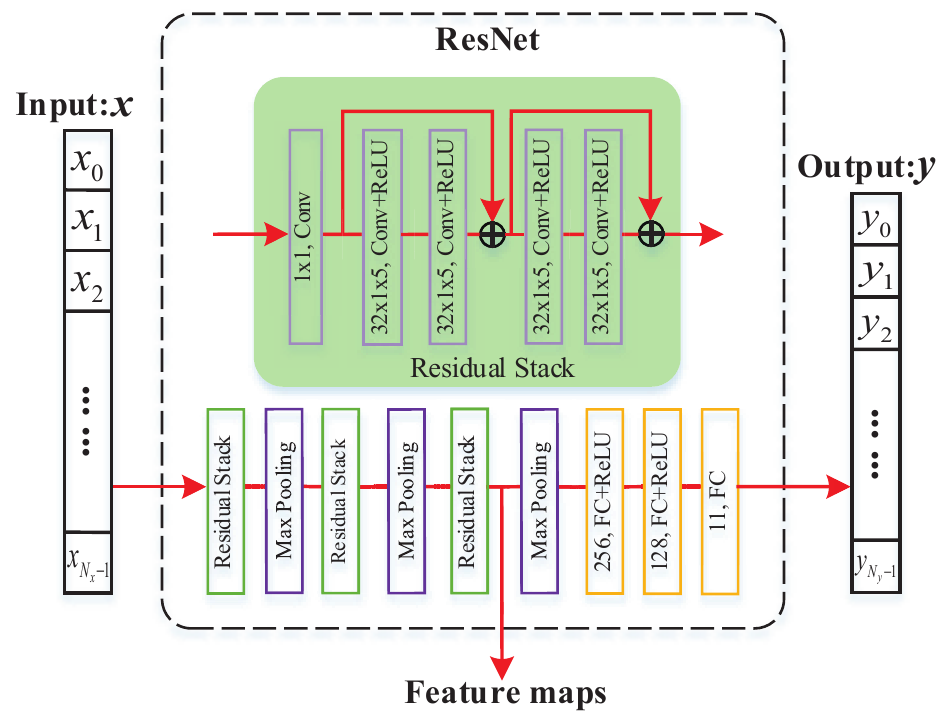}
	\caption{ResNet Structure.}
	\label{fig:3}
\end{figure}

\begin{figure*}[ht]
	\centering
	\includegraphics[scale=0.9]{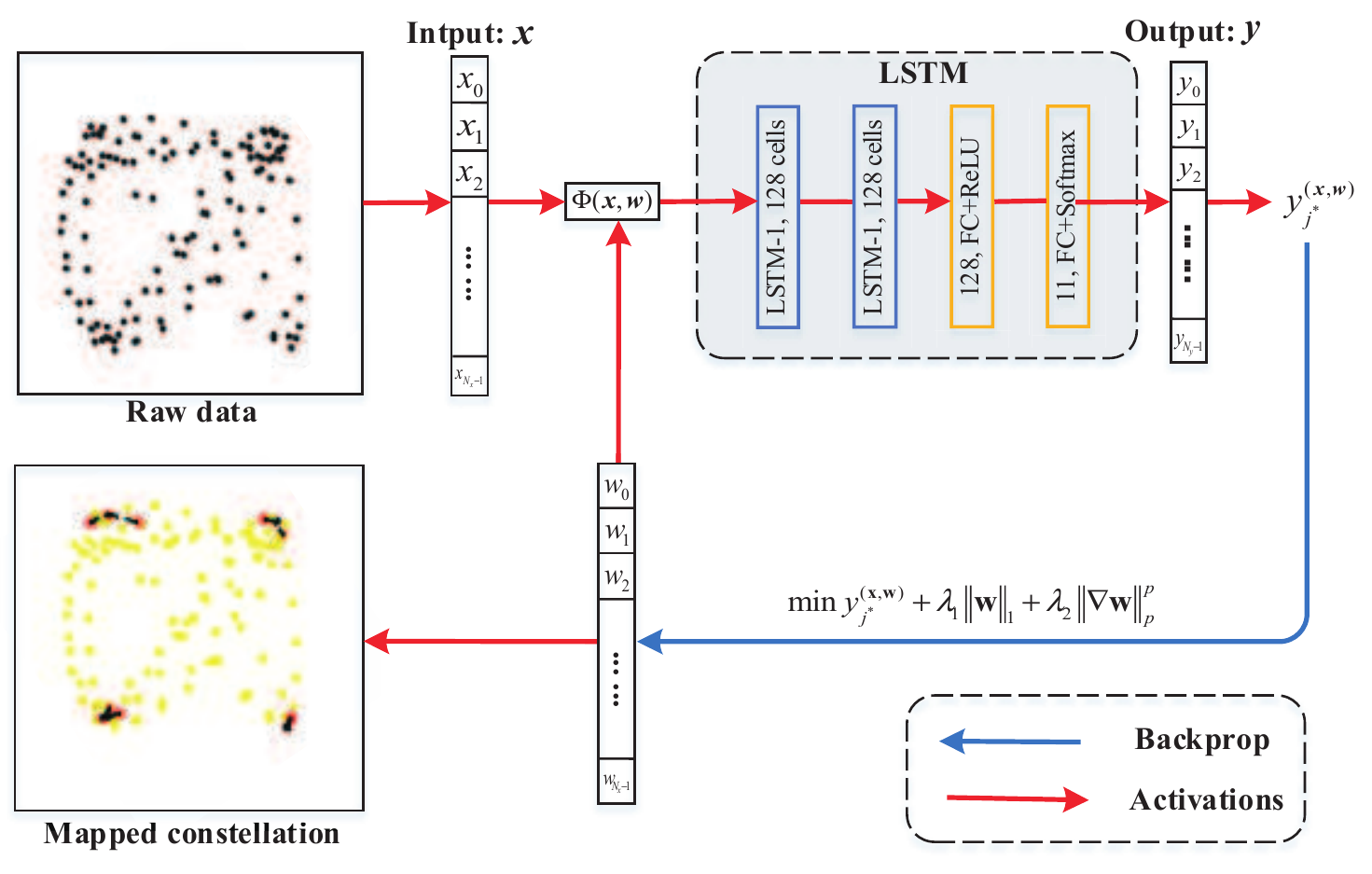}
	\caption{The schematics of visualizing LSTM-based radio modulation classifier.}
	\label{fig:4}
\end{figure*}

\subsection{Visualize CNN-based Classifier}
A CNN-based radio modulation classifier uses different convolutional layers to capture different radio features and feeds them to a fully connected neural network for modulation classification. In general, the deeper convolutional layer captures the higher-level radio feature. Therefore, we adopt the radio features resulted from the last convolutional layer and compute the weight $\boldsymbol{w}$ based on the Grad-CAM \cite{Selvaraju_2017_ICCV} algorithm, as illustrated in Fig.~\ref{fig:2}. Grad-CAM is a widely used for visual explanation in the fields of image recognition. It uses the gradient information flowing into the last convolutional layer and output a heat map visualizing the importance of each neuron for predicting modulation categories.

In CNN-based classifier, each convolutional layer is composed of a series of unique convolutional kernels, where each convolution kernel extracts one radio feature, the so-called feature map. Denote the feature maps generated by the last convolutional layer as $\boldsymbol{F}=\left\{\boldsymbol{f}^{k} | k \in \mathbb{N}_{f}\right\}$, $\mathbb{N}_{f}=\left\{1,2, \cdots, N_{f}\right\}$, where $N_{f}$ is the number of the feature maps and $\boldsymbol{f}^{k}$ is the k-th feature map with width $U$ ($U=1$ in this paper) and height $V.$ Given an input radio signal ${x}$ along with its predicted modulation
category $j^{*}$, Grad-CAM computes a weight $\alpha_{k}^{{j}^{*}}$ which captures the importance of
feature map $\boldsymbol{f}^{k}$ for the predicted category $j^{*}$, as
\begin{eqnarray}
\alpha_{k}^{j^{*}}=\frac{1}{U V} \sum_{u=1}^{U} \sum_{v=1}^{V} \frac{\partial y_{j^{*}}}{\partial f_{u v}^{k}},
\end{eqnarray}
where $f_{u v}^{k}$ refers to the activation value at location $(u, v)$ of the feature map $\boldsymbol{f}^{k}$ and $\frac{\partial y_{{j}^{*}}}{\partial f_{u v}^{k}}$ computes the gradient of the predicted score  $y_{j^{*}}$ with respect to the feature map $\boldsymbol{f}^{k}.$ Note that the score $y_{j^{*}}$ is before the softmax activation function. By summing all weighted feature maps, the class activation vector $\boldsymbol{w}$ is obtained from
\begin{eqnarray}
\boldsymbol{w}=\max \left(\sum_{k=1}^{N_{f}} \alpha_{k}^{j^{*}} \boldsymbol{f}^{k}, 0\right),
\end{eqnarray}
where the max operation keeps features that have a positive influence on the modulation category $j^{*}$. For the sake of better visualization, $\boldsymbol{w}$ is further uniformly normalized to the region $[0,1]$. Note that, the class activation vector  $\boldsymbol{w}$ has the same size as the feature map  $\boldsymbol{f}^{k}$ ($1 \times V$ in this paper), which needs be resized\footnote{In this paper, we adopt the resize() function with the default bilinear interpolation from the OpenCV library https://opencv.org/.} to $1 \times N_{x}$ when $V \neq N_{x}$. In this paper, we evaluate two typical CNN-based radio modulation classifiers, i.e., the LeNet-based classifier \cite{10.1007/978-3-319-44188-7_16} and the ResNet-based classifier \cite{ramjee2019fast}.

\subsubsection{LeNet-based Classifier}
A LeNet-based classifier was first used to successfully classify radio modulation categories in \cite{10.1007/978-3-319-44188-7_16} in 2016. The evaluated LeNet model is composed of two convolutional layers and three fully connected layers and its detailed structure is shown in Fig.~\ref{fig:2}. Then, we use the feature maps from the second convolutional layer for visualization whose total number is $N_{f}=80$ and compute the class activation vector  $\boldsymbol{w}$ based on equation (3).

\subsubsection{ResNet-based Classifier}
A ResNet-based classifier was used to further improve the modulation classification accuracy in \cite{ramjee2019fast}. The ResNet model is composed of three residual stacks and three fully connected layers, where each residual stack contains 5 convolutional layers and one max-pooling layer. However, the max-pooling layer inside each residual stack reduces the dimensionality of the feature maps from the last convolutional layer. In order to obtain the original feature maps, we modify the ResNet by moving the max-pooling layer out of the residual stack, as shown in Fig.~\ref{fig:3}. Then, each residual stack is followed by a max-pooling layer, and we compute the class activation vector $\boldsymbol{w}$ based on the feature maps from the third residual stack.

\subsection{Visualize LSTM-based Classifier}
We study a state-of-art LSTM-based classifier which achieves similar classifying accuracy as the ResNet-based classifier. It is composed of two LSTM layers and two fully connected layers \cite{rajendran2018deep} as shown in Fig.~\ref{fig:4}. In each LSTM layer, a series of LSTM cells are consecutively connected to capture the time-domain feature of continuously-valued radio samples. Therefore, we cannot simply capture the feature maps after the last LSTM layer for visualization. Instead, we dynamically optimize a class activation vector $\boldsymbol{w}$ by masking the radio samples $\boldsymbol{x}$ and minimizing the predicted probability  $y_{j^{*}}$ of the target modulation category.

The structure for visualizing the LSTM-based classifier as illustrated in Fig.~\ref{fig:4}. Firstly, we define a mask function $\Phi(\boldsymbol{x}, \boldsymbol{w})$ \cite{Fong_2017_ICCV} as
\begin{eqnarray}
\Phi(\boldsymbol{x}, \boldsymbol{w})=(\boldsymbol{1}-\boldsymbol{w}) \odot \boldsymbol{x}+\xi\boldsymbol{w},
\end{eqnarray}
where $\boldsymbol{1}$ is an all-ones vector, $\bigodot$ is the Hadamard product and $\xi$ is a constant deletion value to compensate the masked input. Then, we feed the masked samples into the LSTM-based classifier and obtain a new prediction,  {$\boldsymbol{y} = \pi(\Phi(\boldsymbol{x}, \boldsymbol{w}))$. For brevity, we denote the predicted probability for the $j^{*}-$th modulation category as $y_{j^*}^{(\boldsymbol{x}, \boldsymbol{w})}= \pi_{j^*}(\Phi(\boldsymbol{x}, \boldsymbol{w}))$.} Given an LSTM-based classifier, we aim to solve for a class activation vector $\boldsymbol{w}$ that minimizes the following objective function \cite{Fong_2017_ICCV}
\begin{eqnarray}
 {\arg \min _{\boldsymbol{w}} y_{j^*}^{(\boldsymbol{x}, \boldsymbol{w})} +\lambda_{1}\|\boldsymbol{w}\|_{1}+\lambda_{2}\|\nabla \boldsymbol{w}\|_{p}^{p},
\label{eq:lstm_visual}}
\end{eqnarray}
where $\|\cdot\|_{p}$ is $p-$th norm operation, and $\lambda_{1}$ and $\lambda_{2}$ are two regulation parameters. The $L-$1 regulation term $\lambda_{1}\|\boldsymbol{w}\|_{1}$ generates more zero values in  $\boldsymbol{w}$ and the total variation (TV) norm regulation term $\lambda_{2}\|\nabla \boldsymbol{w}\|_{p}^{p}$ reduces the artifact in the visualization \cite{Fong_2017_ICCV}. The visualization performance of LSTM-based classifiers greatly depends on parameters $p$, $\lambda_{1}$ and $\lambda_{2}$, whose settings are numerically studied in the next section. Note that the visualization structure presented in Fig.~\ref{fig:4} is model-agnostic, which can also be used to visualize CNN-based classifiers by replacing the LSTM model with the CNN model. However, in practice, it fails to converge in optimizing (\ref{eq:lstm_visual}) when we apply it to ResNet-based classifier.

Once the class activation vector $\boldsymbol{w}$ is obtained, we are ready to visualize the deep learning-based classifier with the method presented in Sec.~\ref{4.1}. In the next section, we extensively visualize and study both CNN-based and LSTM-based classifiers.

\section{Numerical Results}\label{5}
In this section, we visualize deep learning-based modulation classifiers based on an open dataset, RadioML2016.10a \cite{grcon}. The dataset considers random processes for center frequency offset, sample rate offset, additive white Gaussian Factors such as noise, multi-path, and fading. And the digital modulations are pulse shaped using 8 samples per symbol. The dataset contains modulated $(I, Q)$ radio signals under 11 modulation categories and different signal-to-noise (SNR) ratios. Specifically, there are 8 digital modulation categories (BPSK, QPSK, 8PSK, 16QAM, 64QAM, GFSK, CPFSK, and PAM4) and 3 analog modulation categories (WB-FM, AM-SSB, and AM-DSB), each of which contains 1000 modulated signal samples with length 128 per SNR. In this paper, we evaluate different classifiers based on 110,000 signal samples, whose SNRs range from 0 dB to 18 dB with a step size of 2dB. The dataset is randomly split into training, validation, and test subsets with sizes of 88,000, 11,000, and 11,000, respectively. By using the PyTorch platform \cite{ramjee2019fast}, we implement all three modulation classifiers presented in Sec.~\ref{4} and successfully train their deep learning models with parameters given in Table~\ref{t_trainingParameters}. The obtained modulation classification accuracies for classifiers based on LeNet, ResNet, and LSTM are around 86\%, 92\% and 92\% at 18dB SNR, which agree with the results reported in \cite{10.1007/978-3-319-44188-7_16,ramjee2019fast,rajendran2018deep}.

\begin{table} 
	\centering\small
	\begin{threeparttable}
		\caption{Training parameters for different deep learning-based radio modulation classifiers}
		\setlength{\tabcolsep}{0.48mm}{
		\begin{tabular}{lcccccc} 
			\toprule 
		    &Bath  & Learning  & \multirow{2}{*}{Epoch } & \multirow{2}{*}{Dropout } & \multirow{2}{*}{Optimizer } & \multirow{2}{*}{Accuracy }  \\
            & Size &  Rate &  &  &  &   \\
			\midrule 
			LeNet    &   128 & 0.001 & 150 & 0.5 & Adam & 86\%  \\
			\midrule 
			ResNet      &   128 & 0.001 & 150 & 0.25 & Adam & 92\%  \\
			\midrule 
			LSTM       &   128 & 0.001 & 150 & 0.5 & Adam & 92\%  \\
			\bottomrule 
		\end{tabular}}
	\end{threeparttable}
\label{t_trainingParameters}
\end{table}

\subsection{Visualization Parameters}

\begin{figure}
    \centering
    \includegraphics[scale=1]{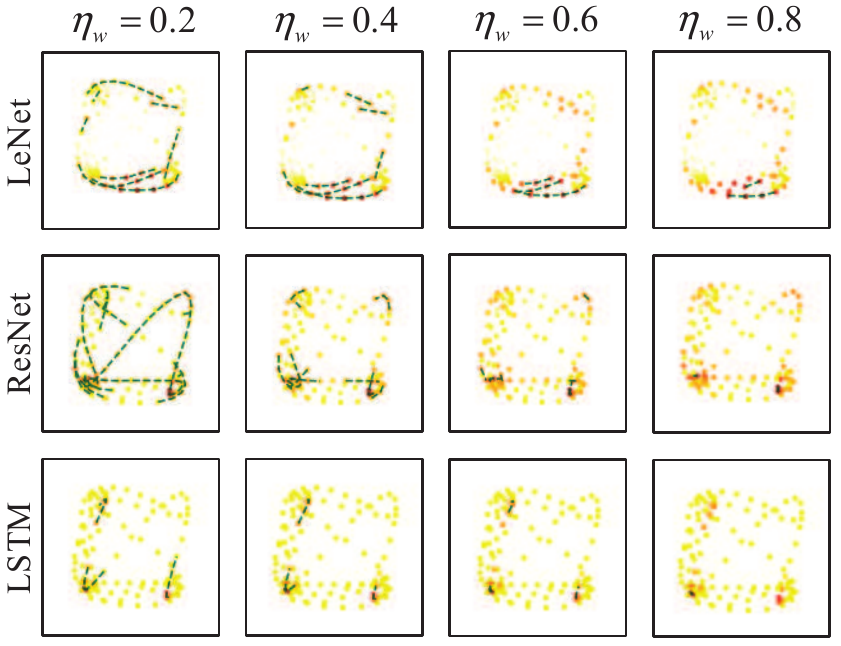}
    \caption{QPSK modulation visualization under different thresholds $\eta_{w}$ for LeNet-based, ResNet-based, and LSTM-based radio modulation classifiers.}
    \label{fig:5}
\end{figure}
In Fig.~\ref{fig:5}, we visualize classifiers' time-domain features under different weight thresholds $\eta_{w}$. A small threshold $\eta_{w}$ includes too much signal features to discriminate their significance, i.e., $\eta_{w}=0.2$. On the other hand, a large threshold $\eta_{w}$ filters most of the radio samples and fails to capture the time-domain feature, i.e., $\eta_{w}=0.8$. Therefore, we set $\eta_{w}=0.4$ in the rest of this paper.


\begin{figure}
    \centering
    \includegraphics[scale=1]{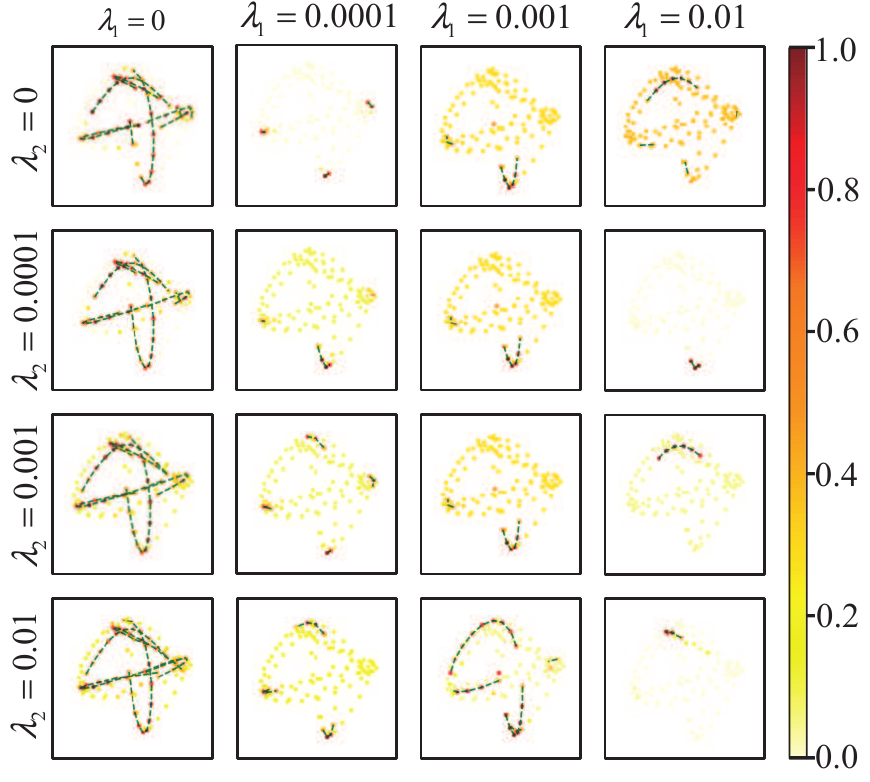}
    \caption{QPSK modulation visualization under the L-1 regulation parameter $\lambda_{1}$ and the TV
        regulation parameter $\lambda_{2}$ for the LSTM-based classifier.}
    \label{fig:6}
\end{figure}
In Fig.~\ref{fig:6}, we study visualization performance of the LSTM-based classifier under different regulation parameters. Mathematically, by setting $\lambda_{1}=\lambda_{2}=0$, the predicted probability $y_{j^{*}}^{(\boldsymbol{x}, \boldsymbol{w})}$ is minimized. However, it may result in artifacts without physical insight. A greater value of the L-1 regulation parameter $\lambda_{1}$ results in more zero-value elements in the optimized $\boldsymbol{w}$, e.g., few red-colored sample points when $\lambda_{1}=0.01$. On the other hand, when $\lambda_{1}=0$, $\boldsymbol{w}$ is not minimized and many sample points are red-colored. In order to highlight these most important sample points, we set $\lambda_{1}=0.0001$ in the following simulations. The TV regulation parameter $\lambda_{2}$ regulates the difference between two consecutive elements of $\boldsymbol{w}$. Increasing the value of $\lambda_{2}$ causes the connected points to be more consecutive, e.g., when $\lambda_{1}=0.0001$ and $\lambda_{2}=0.01$. As a compromise between the reduction of the predicted probability $y_{j^{*}}^{(\boldsymbol{x}, \boldsymbol{w})}$ and the smoothness for physical explanation, we set $\lambda_{2}=0.001$ in the following simulations.

\begin{figure}
    \centering
    \includegraphics[scale=1]{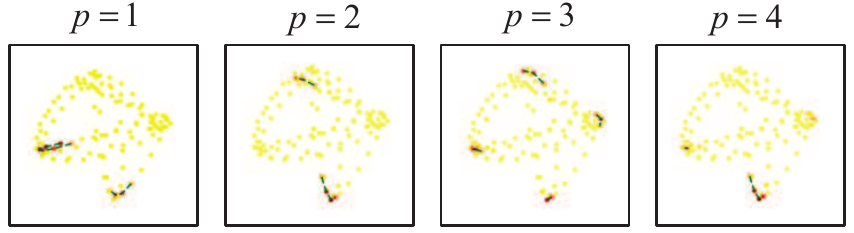}
    \caption{QPSK modulation visualization under different orders of the norm $p$ for the LSTM-based classifier.}
    \label{fig:7}
\end{figure}
In Fig.~\ref{fig:7}, we illustrate the visualizations under different orders of the norm $p$. A smaller $p$ pursues the smoothness of $\boldsymbol{w}$. For example, only two consecutive segments are connected when $p=1$. However, it also results in greater value of the TV regulation term $\lambda_{2}\|\nabla \boldsymbol{w}\|_{p}^{p}$ in the objective presented in equation (5), causing difficulty in reducing the predicted probability $y_{j^{*}}^{(\boldsymbol{x}, \boldsymbol{w})}$. As a compromise between smoothness and accuracy, we set $p=3$ in the rest of this paper.

\begin{figure}
    \centering
    \includegraphics[scale=1.1]{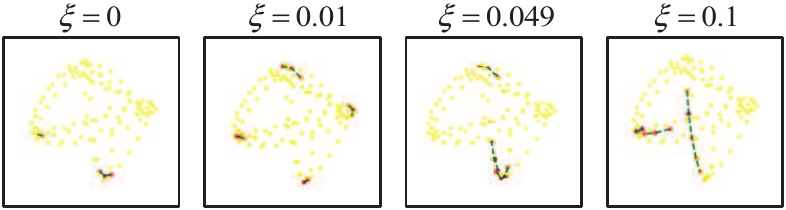}
    \caption{QPSK modulation visualization under different deletion values $\xi$.}
    \label{fig:8}
\end{figure}
In Fig.~\ref{fig:8}, we investigate the classifier visualization under different deletion values $\xi$. When $\xi=0$, the masked input is unnatural since those significant elements in the input are simply removed. We also evaluate the deletion value as the mean of input signals, i.e., $\xi=0.049$ for the considered radio signals. A larger deletion value will increase the difficulty in converging to optimal. Hence, we set $\xi=0.01$ in the following evaluations and summarize all those parameters in Table~\ref{t_parameters}.

\begin{figure*}[]
    \centering
    \includegraphics[width=1\textwidth]{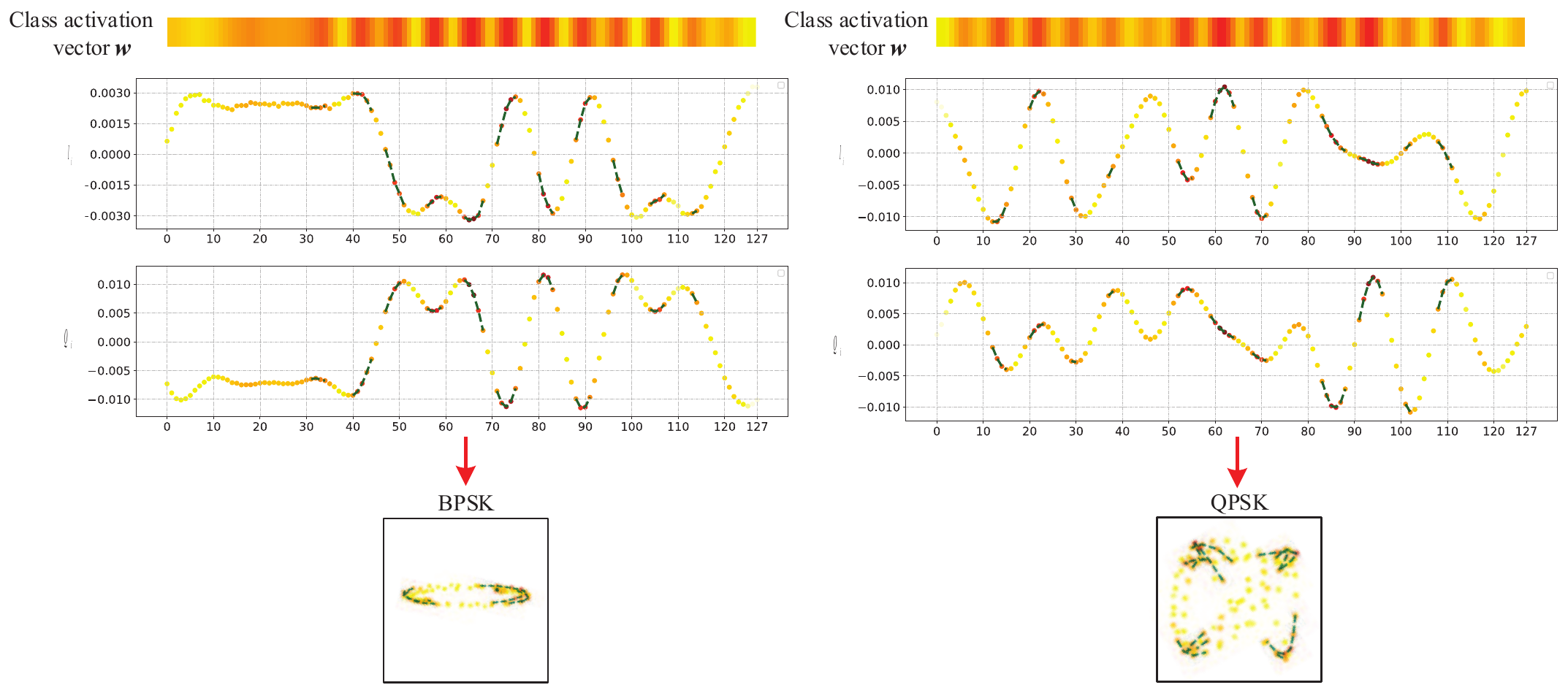}
    \caption{Visualizations of radio features from the class activation vector  $\boldsymbol{w}$.}
    \label{fig:91}
\end{figure*}
In Fig.~\ref{fig:91}, we illustrate how to visualize radio features in both the time-domain signals and the constellation from the obtained class activation vector $\boldsymbol{w}$. Specifically, both BPSK and QPSK modulation examples from the ResNet-based modulation classifier are visualized. The $\boldsymbol{w}$ has been resized to the scale of $1 \times N_{x}$. In the time domain, the modulated in-phase signal $I_{i}$ and quadrature-phase signal $Q_{i}$ are mapped with the same weight $w_i$. To better illustrate radio features, we further convert the mapped time-domain signals $(I_{i},Q_{i})$ into mapped constellations, as illustrated in Fig.~\ref{fig:91}.


\begin{table} 
	\centering
		\caption{Visualization parameters for LSTM-based radio modulation classifier}
		\begin{tabular}{r|l} 
            parameter & value  \\ \hline
			weight threshold $ \eta_{w}$&0.4\\
			$L-1$ regulation parameter $\lambda_{1}$ &0.0001\\
			TV norm regulation parameter $\lambda_{2}$&0.001\\
			deletion value $\xi$&0.01\\
		\end{tabular}
\label{t_parameters}
\end{table}

\subsection{Visualization with Different Input Formats}

\begin{figure*}
    \centering
    \includegraphics[width=1\textwidth]{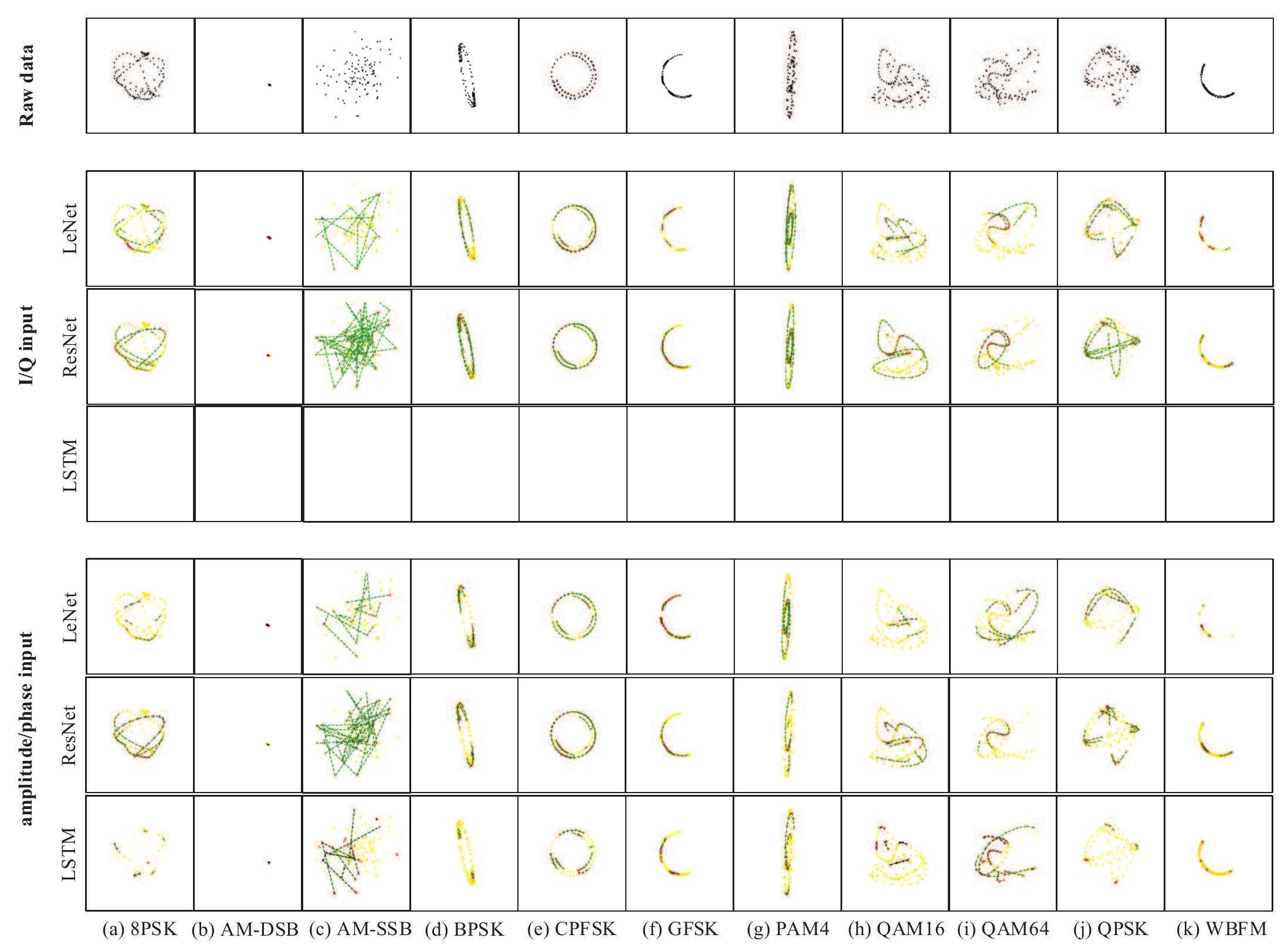}
    \caption{Visualization of different deep learning-based radio modulation classifiers under different input formats.}
    \label{fig:9}
\end{figure*}
In Fig.~\ref{fig:9}, we visualize three deep learning-based radio modulation classifiers under different input formats. The grey raw data illustrate the constellations of original 11 types of modulated signals, whose mapped constellations from different modulation classifiers are provided below. As shown in Fig.~\ref{fig:9}, different modulations are better differentiated after visualizations. For each signal sample, we feed the classifiers with both I/Q input $x_{i}=\left(I_{i}, Q_{i}\right)$ and amplitude/phase input $x_{i}=\left(\mathrm{A}_{i}, \phi_{i}\right)$. As reported in \cite{rajendran2018deep}, the LSTM-based classifier fails to classify modulation categories with I/Q inputs, whose corresponding visualization is not available. Visualizations of the CNN-based classifiers are similar, which are insensitive to the input formats. For each of the 11 modulation categories, both the LeNet-based classifier and the ResNet-based classifier capture almost the same radio signal features under both I/Q inputs and amplitude/phase inputs. Considering the amplitude/phase input, all three classifiers capture similar radio features for different modulation categories except that there are fewer connected radio sample points for the LSTM-based classifier.

\begin{figure}
    \centering
    \includegraphics[scale=1]{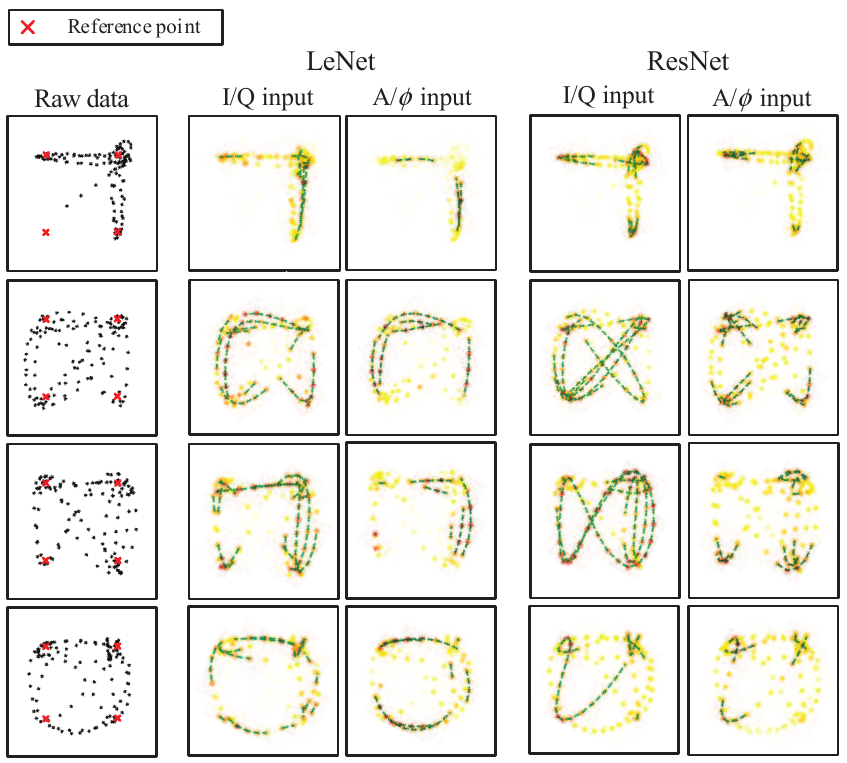}
    \caption{Visualization of LeNet-based and ResNet-based classifiers with different input formats.}
    \label{fig:10}
\end{figure}

\begin{figure}
	\centering
	\includegraphics[scale=0.62]{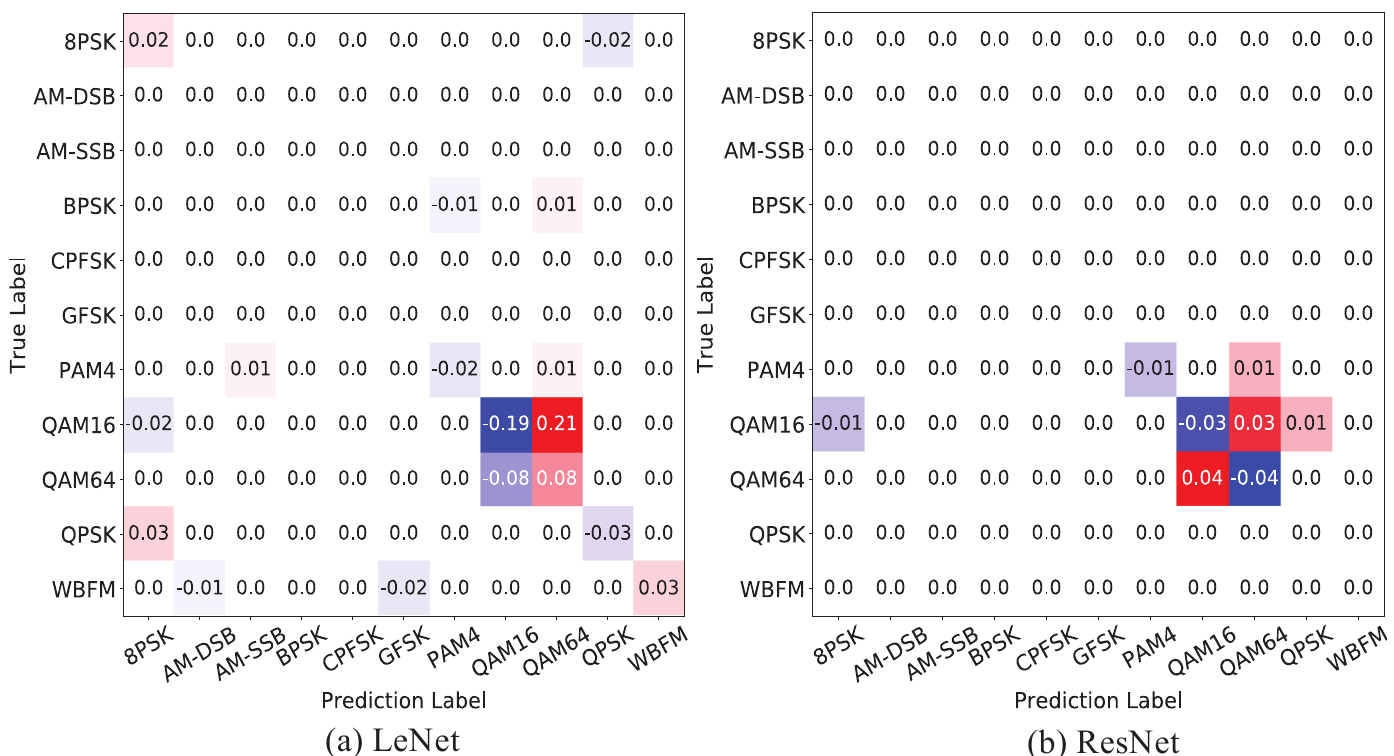}
	\caption{Relative confusion matrix of classifiers under different input formats: (a) LeNet-based classifier
		and (b) ResNet-based classifier.}
	\label{fig:11}
\end{figure}

In Fig.~\ref{fig:10}, we further study the LeNet-based and ResNet-based classifiers with different input formats via QPSK modulation examples. For both CNN-based modulation classifiers, the class activation vector $\boldsymbol{w}$ has greater weights when using the I/Q inputs than using the amplitude/phase inputs, resulting more connected sample points. Interestingly, there is a little improvement in the classification accuracy (less than 1\%) when using I/Q as the input, as the relative confusion matrix shown in Fig.~\ref{fig:11}. Here, the relative confusion matrix is obtained by subtracting the confusion matrix obtained with the amplitude/phase inputs from the one obtained with the I/Q inputs. Therefore, I/Q inputs are preferred for the CNN-based classifiers which capture more meaningful radio features.

\subsection{Visualization for Different Classifiers}
\begin{figure*}
    \centering
    \includegraphics[scale=1]{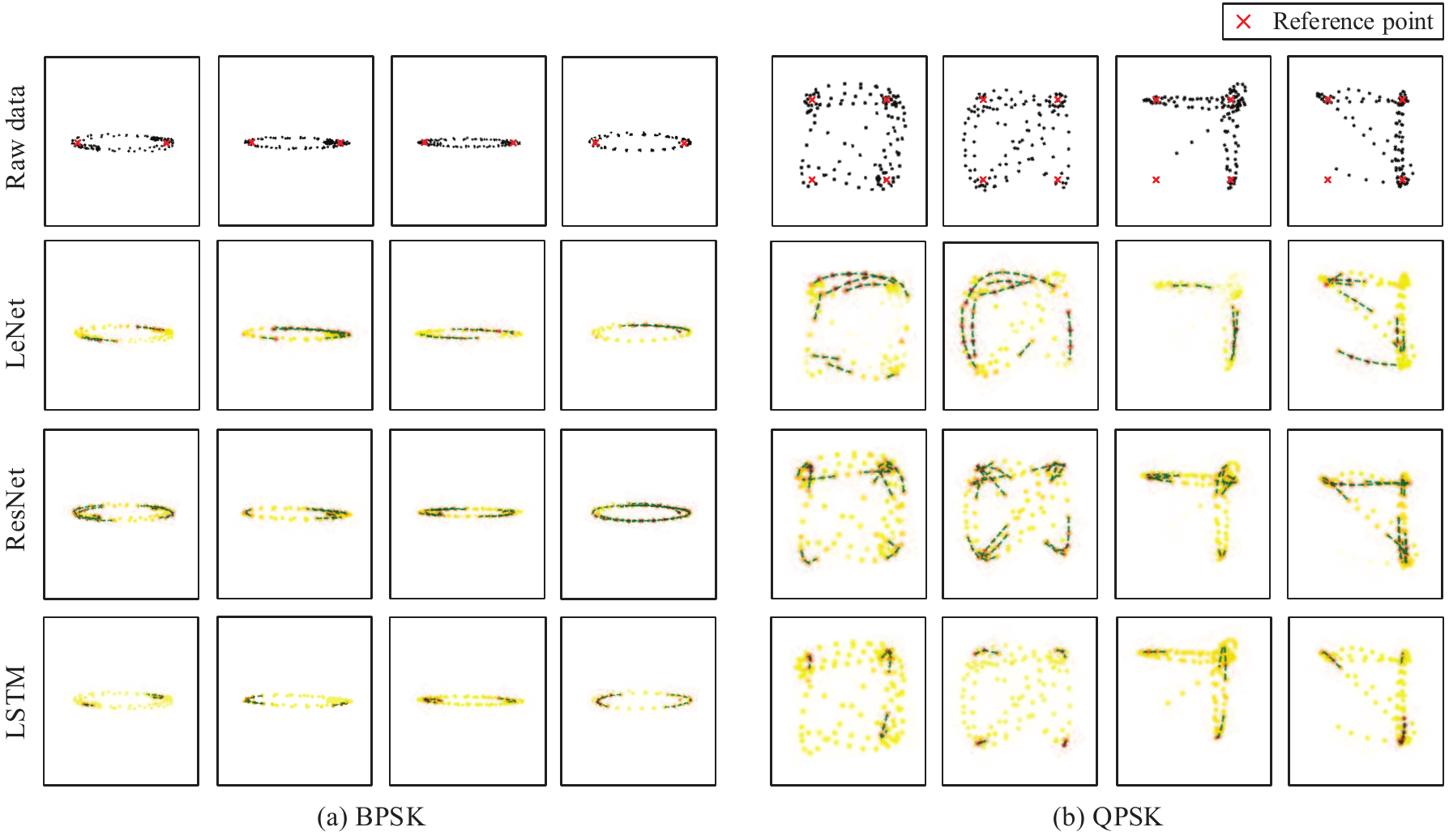}
    \caption{Visualization of different deep learning-based radio modulation classifiers (a) BPSK and (b) QPSK modulations.}
    \label{fig:12}
\end{figure*}
In Fig.~\ref{fig:12}, we compare the visualizations of all three classifiers via BPSK and QPSK modulations with amplitude/phase inputs. For BPSK modulation, there are two reference points and each reference point represents one symbol. During data
transmission, the effective symbol stochastically alternates between these two reference points whose transition process is sampled and represented by the radio signal samples shown in Fig.~\ref{fig:12} (a). Both CNN-based classifiers capture the transition process to classify BPSK modulation, where the LeNet-based classifier captures the transition between the reference points and the ResNet-based classifier prefers the transition around the reference points. On the other hand, the LSTM-based classifier discriminates BPSK from other modulation categories only based on those sample points close to the reference points, which is similar to the knowledge of human experts. In Fig.~\ref{fig:12} (b), we observe similar behaviors of three classifiers via QPSK examples. The QPSK modulation owns four reference points whose visiting times are randomly depending on transmitted data content. In this paper, each radio sample contains only 128 sample points and the visit to every reference point cannot be guaranteed. However, all four QPSK radio samples are successfully discriminated by three classifiers.

\begin{figure*}
    \centering
    \includegraphics[scale=1]{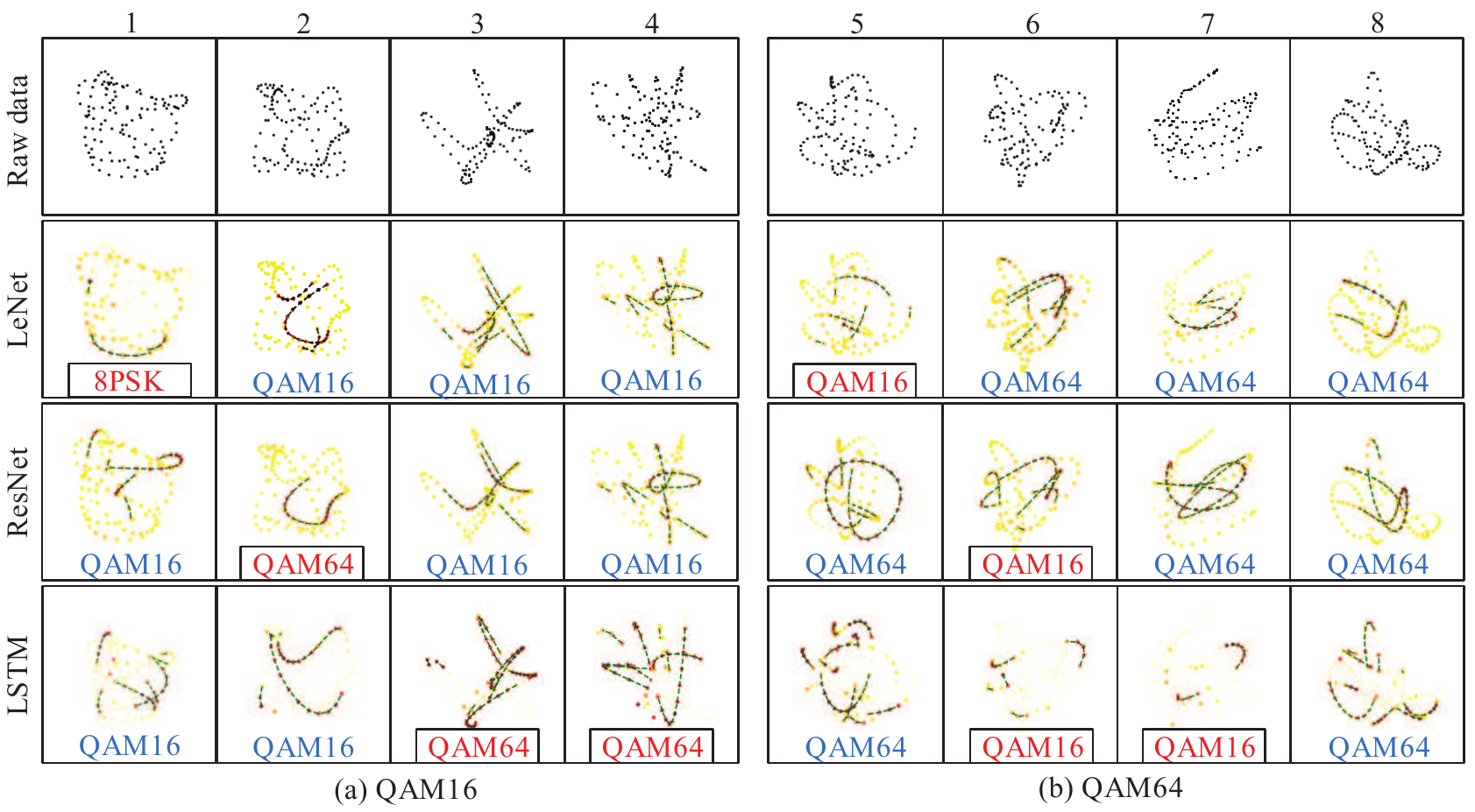}
    \caption{Visualization of different deep learning-based radio modulation classifiers with misclassified radio samples: (a) QAM16
        modulation and (b) QAM64 modulation.}
    \label{fig:13}
\end{figure*}
\begin{figure*}
    \centering
    \includegraphics[scale=1.35]{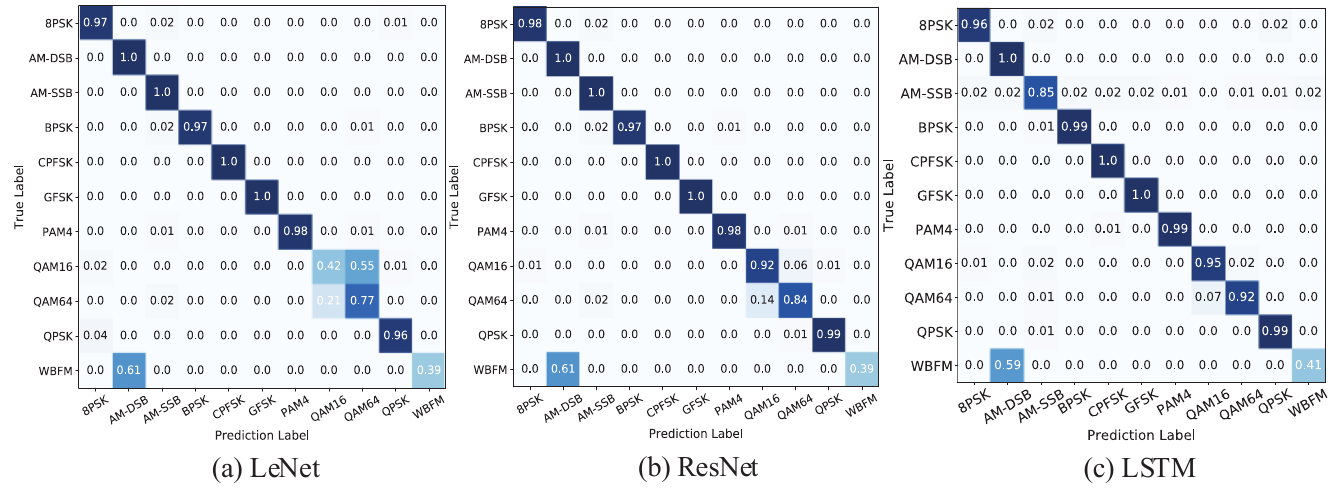}
    \caption{Confusion matrix diagrams for different deep learning-based radio modulation classifiers: (a) LeNet-based
        classifier, (b) ResNet-based classifier, and (c) LSTM-based classifier.}
    \label{fig:14}
\end{figure*}
In Fig.~\ref{fig:13}, we further visualize three deep learning-based classifiers when the radio samples are misclassified. Specifically, eight radio samples from QAM16 and QAM64 modulations are investigated and their corresponding predicted modulation categories are marked in blue (red) labels for successful (failed) predictions. In general, it is difficult for all three classifiers to discriminate QAM16 and QAM64, as shown via the confusion matrix in Fig.~\ref{fig:14}. For example, the first radio sample belongs to the QAM16 modulation category, which is successfully predicted by both ResNet-based and LSTM-based classifiers. However, the LeNet-based classifier captures circularly connected feature and misclassifies it as the 8PSK modulation. The classification result depends on the radio features captured by each classifier, which greatly depends on the specific radio sample. Although the average prediction accuracy of the LeNet-based classifier is smaller than the Resnet-based and LSTM-based classifiers, it successfully classified the QAM64 radio sample shown in the sixth column while the other two classifiers failed.

\subsection{Visualization on Short Samples}

\begin{figure*}
    \centering
    \includegraphics[scale=1]{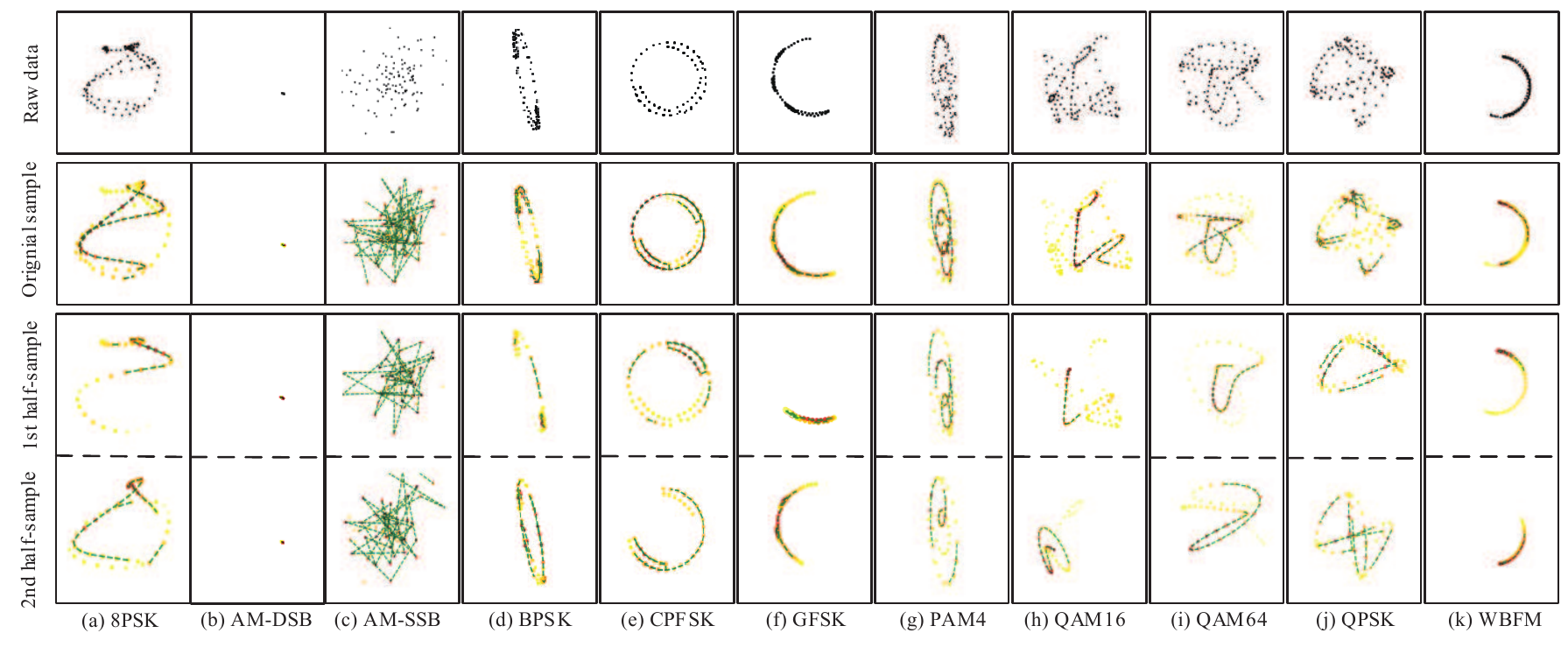}
    \caption{Visualization of different deep learning-based radio modulation classifiers before and after splitting the 128-point radio samples.}
    \label{fig:15}
\end{figure*}
\begin{figure*}
    \centering
    \includegraphics[scale=1]{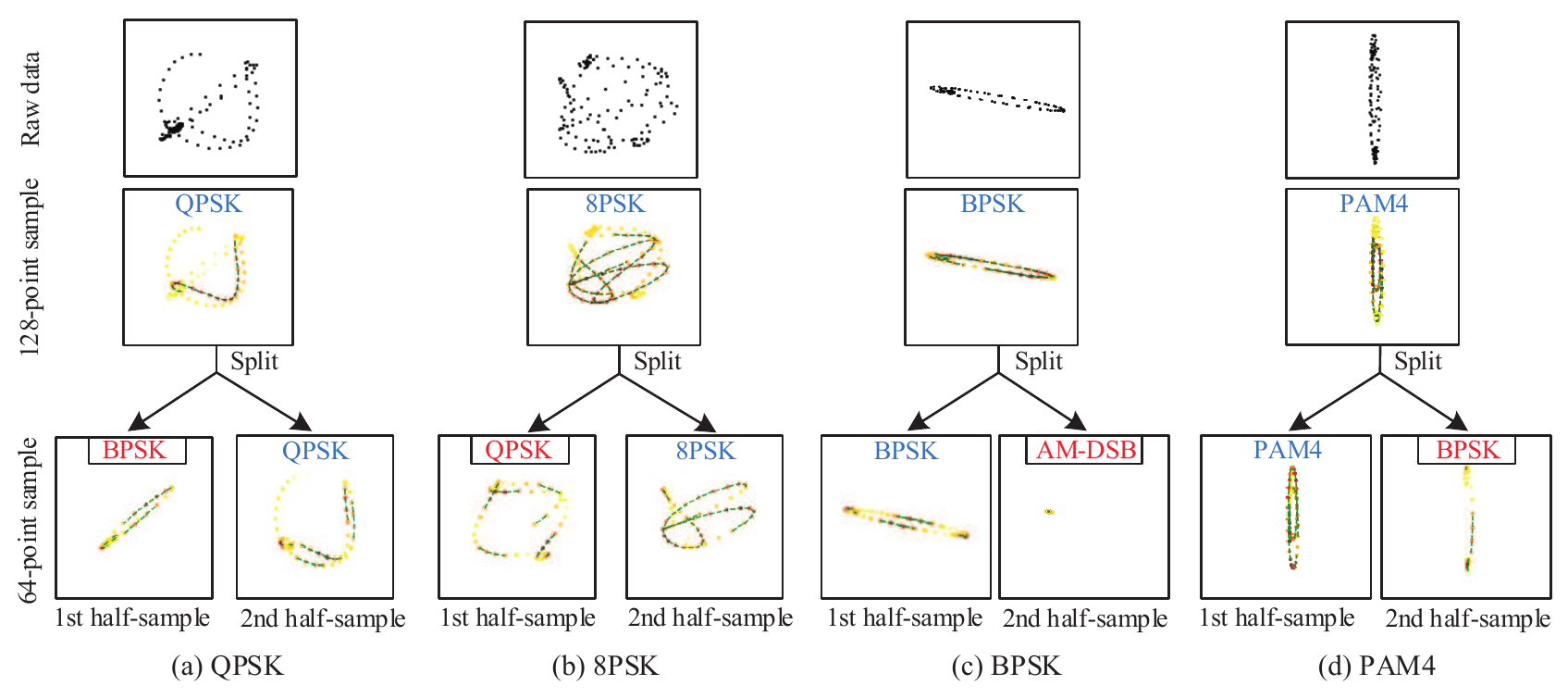}
    \caption{Classifier visualization when the 64-point radio samples are misclassified: (a) QPSK, (b) 8PSK,
        (c) BPSK, and (d) PAM4.}
    \label{fig:16}
\end{figure*}
In Fig.~\ref{fig:15}, we visualize the deep learning-based modulation classifier with fewer sample points. Specifically, we split each 128-point radio sample into two 64-point signal samples, retrain the ResNet model as before, and obtain a ResNet-based classifier with an accuracy of 90\%, which is a 2\% reduction from the original 128-point case. As shown in Fig.~\ref{fig:15}, the radio features of each original 128-point sample are equally distributed and captured by both the 64-point samples. Nevertheless, there still exists a short radio sample that cannot capture the whole features of the radio signals, as illustrated in Fig.~\ref{fig:16}. Taking the QPSK modulation in Fig.~\ref{fig:16} (a) as an example, its modulation category is successfully classified based on the original 128-point radio sample. After the splitting operation, the first 64-point sample distributes between two out of the four reference points, whose captured features are similar to the BPSK modulation, resulting a misclassification. Meanwhile, the second 64-point sample keeps most of the original radio features and is successfully classified as QPSK modulation. Similar results are visualized for the 8PSK, BPSK, and PAM4 modulations shown in Fig.~\ref{fig:16} (b), (c), and (d), respectively. By visualizing the extracted radio features, we show that the classification accuracy greatly depends on the contents carried by radio signals, which explains why the classification accuracy decreases with shorter samples \cite{8936957}.

\section{Conclusion}\label{6}
In this paper, we proposed a visualization technique to study the radio features extracted by different deep learning-based radio modulation classifiers, i.e., the CNN-based classifier and the LSTM-based classifier. By studying radio signals under both I/Q and amplitude/phase formats, we show that CNN-based classifiers are insensitive to the input formats and capture similar radio features. Specifically, the LeNet-based classifier captures the transitions between modulation reference points, while the ResNet-based classifier prefers to capture the transitions around modulation reference points. In comparison, the LSTM-based classifier discriminates different modulation categories based on sampling points close to the reference points, which is similar to the knowledge of human experts. We further visualized the ResNet-based classifier under the cases of shorter radio samples. We show that the radio features extracted by the deep learning-based classifier greatly depend on the  contents carried by radio signals and a short radio sample may lead to misclassification. Our proposed visualization technique in this paper is general and can be applied to any other CNN-based or LSTM-based radio modulation classifiers. 

%
%

\ifCLASSOPTIONcaptionsoff
  \newpage
\fi



%
%
%
\bibliographystyle{IEEEtran}

%

%
%
%




\end{document}